\documentclass{isprs} 
\usepackage{graphicx}
\usepackage{url}
\usepackage{setspace}
\usepackage{geometry} 
\usepackage{epstopdf}
\usepackage[labelsep=period]{caption}  
\usepackage[british]{babel} 
\usepackage[hang]{footmisc}
\usepackage{float}
\usepackage{subfigure}
\usepackage{subcaption}
\usepackage{algorithm} 
\usepackage{algpseudocode}
\usepackage{graphicx}
\usepackage{changepage}
\usepackage{booktabs}
\usepackage{wrapfig}
\usepackage{array}
\usepackage{diagbox}
\usepackage{tabularx}
\usepackage{url}
\usepackage{lineno}
\usepackage{xcolor}

\emergencystretch=\maxdimen
\begin{document}

\title{Building-PCC: Building Point Cloud Completion Benchmarks}
\date{}


\author{
Weixiao Gao\textsuperscript{1}\thanks{Corresponding author}, Ravi Peters\textsuperscript{2}, Jantien Stoter\textsuperscript{1} }



\address{
	\textsuperscript{1 }Delft University of Technology, The Netherlands - (w.gao-1,j.e.stoter)@tudelft.nl\\
	\textsuperscript{2 }3DGI, Zoetermeer, The Netherlands - ravi.peters@3dgi.nl\\
}




\abstract{
With the rapid advancement of 3D sensing technologies, obtaining 3D shape information of objects has become increasingly convenient. Lidar technology, with its capability to accurately capture the 3D information of objects at long distances, has been widely applied in the collection of 3D data in urban scenes. However, the collected point cloud data often exhibit incompleteness due to factors such as occlusion, signal absorption, and specular reflection. This paper explores the application of point cloud completion technologies in processing these incomplete data and establishes a new real-world benchmark Building-PCC dataset, to evaluate the performance of existing deep learning methods in the task of urban building point cloud completion. Through a comprehensive evaluation of different methods, we analyze the key challenges faced in building point cloud completion, aiming to promote innovation in the field of 3D geoinformation applications. Our source code is available at \url{https://github.com/tudelft3d/Building-PCC-Building-Point-Cloud-Completion-Benchmarks.git}. 
}



\maketitle


\section{Introduction}\label{sec:Introduction}
The rapid advancement of 3D sensing technology has simplified acquiring 3D object information, making data acquisition increasingly convenient. LiDAR, compared with stereo vision and structured light, can capture 3D information accurately from long distances~\cite{giancola2018survey}. For collecting urban data, LiDAR has been widely used~\cite{wang2019survey}, including airborne, terrestrial, and mobile laser scanning. However, these devices still face issues such as occlusion, signal absorption, reflection, and sensor resolution, which result in incomplete point clouds~\cite{fei2022comprehensive}. Thus, completing these point clouds to restore full structures is crucial, especially for applications related to 3d geoinformation, such as 3D reconstruction~\cite{nan2017polyfit}, semantic understanding~\cite{song2017semantic}, change detection~\cite{czerniawski2021automated}, and autonomous driving~\cite{bai2020depthnet}.

\begin{figure}[ht!]
\centering
\subfigure[AHN3]{\label{fig:fig1a}\includegraphics[width=0.48\columnwidth]{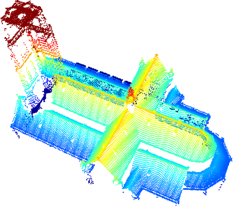}}
\subfigure[AHN4]{\label{fig:fig1b}\includegraphics[width=0.48\columnwidth]{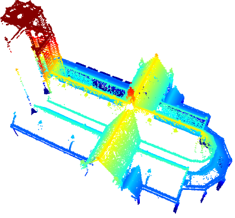}}
\subfigure[LoD2 model]{\label{fig:fig1c}\includegraphics[width=0.48\columnwidth]{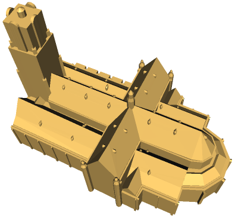}}
\subfigure[Sampled points]{\label{fig:fig1d}\includegraphics[width=0.48\columnwidth]{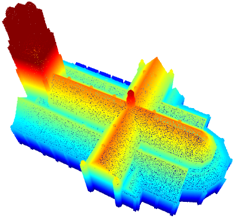}}
\caption{(a) and (b) display partial point clouds captured by airborne LiDAR, corresponding to AHN3 and AHN4, respectively. (c) presents a manually created LoD2 building model, while (d) illustrates point clouds sampled from (c).}
\label{fig:fig1}
\end{figure}

The task of point cloud completion was first proposed in computer vision and graphics fields. Early research inferred the geometric properties of missing areas from observed regions using methods such as surface reconstruction~\cite{berger2014state}, symmetry analysis~\cite{mitra2006partial}, or template matching~\cite{nan2012search}. With the development of deep learning technology, methods based on voxels~\cite{xie2020grnet}, encoders-decoders~\cite{yuan2018pcn}, and transformers~\cite{yu2021pointr,chen2023anchorformer} have been proposed, aiming to recover the complete geometric shape of objects from incomplete data. These learning-based methods can capture higher-dimensional features without relying on prior assumptions, offering significant advantages over traditional methods.

However, despite these advancements, the practical application of these methods predominantly involves testing on CAD model datasets such as ShapeNet~\cite{chang2015shapenet}, revealing a gap in their evaluation within real-world contexts. Existing evaluations in real-world settings often concentrate on specific categories such as cars in datasets like KITTI~\cite{geiger2012we}, lacking comprehensive evaluation for large-scale real-world scenes. To bridge this gap, this paper introduces real-world data benchmarks based on airborne LiDAR point clouds to improve testing scenarios for deep learning-based completion methods.

Airborne point cloud data, known for its high precision, is commonly used for building reconstruction in large-scale urban scenes. For instance, initiatives such as 3D BAG~\cite{peters2022automated} and Helsinki3D~\cite{Helsinki3d} leverage airborne point cloud data as their foundational input, integrating this with GIS-based building footprints to reconstruct building models. These models adhere to the CityGML standards~\cite{kolbe2005citygml}, achieving detailed representations at either Level of Detail 1 (LoD1) or Level of Detail 2 (LoD2) (see Figure~\ref{fig:fig1c}). These models can be applied to various fields~\cite{biljecki2015applications} such as urban planning, energy estimation, fluid dynamics simulation, and city management. However, due to factors such as tree occlusion, glass adsorption, and varying viewpoints, the collected data often have missing parts (see Figure~\ref{fig:fig1a} and~\ref{fig:fig1b}), which critically affect the quality of 3D reconstruction of the building models. Therefore, this research conducts a thorough evaluation of the performance of state-of-the-art, deep learning-based approaches in the task of completing building point clouds.

The objective of this paper is to introduce a new real-world benchmark dataset, namely Building-PCC, specifically designed for the task of building point cloud completion. We collected 50,000 building instances from two cities in the Netherlands, The Hague and Rotterdam, correlating each with two sets of airborne point cloud data, namely AHN3 and AHN4, alongside ground truth data sampled from manually reconstructed 3D building models. Compared to existing datasets, our dataset is based on real-world scenarios and presents a higher level of challenge owing to the intricate and diverse nature of these real-world settings. On this basis, this study comprehensively evaluates the existing deep learning-based 3D completion methods on our dataset. We delve into the critical challenges encountered in building point cloud completion from various perspectives, with the goal of fostering innovation within the realm of 3D geoinformation applications.

\section{Related work}\label{sec:related_work}
This section begins with a comprehensive overview of the datasets frequently employed in 3D point cloud completion tasks. It then proceeds to offer an in-depth exploration of deep learning-based techniques for achieving 3D shape completion.

\begin{figure*}[ht!]
\begin{center}
        \includegraphics[width=1.0\textwidth]{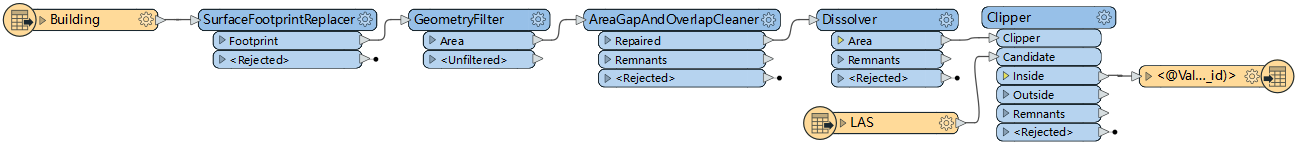}
	\caption{FME workflow for extracting building point cloud instances.}
    \label{fig:fig2}
\end{center}
\end{figure*}

\subsection{Point cloud completion datasets}\label{sec:recons_repair}

Point cloud completion tasks leverage datasets that can be categorized into two primary types: synthetic and real-world scene datasets. Among these, synthetic datasets such as ShapeNet~\cite{chang2015shapenet}, ModelNet~\cite{vishwanath2009modelnet}, and Completion3D~\cite{tchapmi2019topnet} are extensively utilized. These datasets consist of Computer-Aided Design (CAD) models that span a diverse array of object categories, including but not limited to airplanes, ships, furniture, and electronic devices. The generation of complete point clouds within these datasets involves either uniform sampling or Poisson disk sampling techniques applied to the model surfaces, standardizing the maximum resolution at 16,384 sampling points per model~\cite{yuan2018pcn}.  To obtain incomplete point coulds, a process of simulating 2.5D depth images from various viewpoints and subsequently back-projecting these images into the three-dimensional space is employed to achieve sampling.

In the realm of real-world scene datasets, the KITTI~\cite{geiger2012we} dataset stands out as a primary resource. Acquired using Mobile Laser Scanning (MLS) technology across actual road environments, KITTI was initially crafted for the purpose of autonomous driving research. It encompasses 22 point cloud sequences and 11 semantically annotated categories. Within the scope of point cloud completion tasks, the vehicle category frequently serves as the chosen subject for testing. The inherent sparsity of point clouds within the KITTI dataset often results in the geometric structure of the captured object being incomplete. As a consequence, a common approach among researchers is to initially train their models on synthetic datasets featuring car models. Subsequently, they utilize the incomplete vehicle point clouds from the KITTI dataset, performing uniform sampling to create the requisite datasets for further training and predictive analysis.

It is important to highlight that all datasets undergo a normalization process before being fed into the point cloud completion network. This is a crucial step designed to mitigate the adverse effects of outlier samples. Specifically for KITTI data, each vehicle is marked with a bounding box that provides a reference coordinate system for normalization. While most test datasets, aside from KITTI car datasets, consist of synthetic data, they often fall short in capturing the diversity found in real-world settings, particularly within urban scenes. To address this shortfall, this research introduces a comprehensive airborne point cloud completion benchmark dataset tailored for buildings, thereby aiming to bridge this notable gap.

\subsection{3D shape completion}\label{sec:topo_repair}
In the domain of 3D shape completion, methods leveraging deep learning are primarily categorized into three strategic approaches: point cloud-based, multimodality-based, and those employing Generative Adversarial Networks (GANs).

The point cloud-based completion approach is intricately divided into three distinct methodologies: 3D convolution-based methods, encoder-decoder frameworks, and anchor-based strategies. Techniques rooted in 3D convolution typically require the construction of voxel grids or distance fields~\cite{xie2020grnet}, offering high accuracy at the cost of significant computational resources and challenges in capturing the intricate details of objects. Encoder-decoder-based methods~\cite{yuan2018pcn} adeptly extract both global and local 3D features from the point clouds using the encoder, while the decoder refines the prediction of the object's complete point cloud from a coarse to a fine resolution. The anchor-based approach~\cite{chen2023anchorformer,yu2023adapointr}, on the other hand, determines the positions of anchor points by learning the local features of the input, followed by an upsampling process that leverages these anchor points to reconstruct the object's full shape, integrating both local and global features. This method may also incorporate attention mechanisms to further boost the performance of the network. This research is dedicated to examining the performance of these methods in achieving accurate completion of building point clouds.

Multimodality-based methods~\cite{zhang2021view} leverage image data from single or multiple views combined with partial point clouds for completion. These images can provide auxiliary information about the global shape of the object. Notably, such images usually have a pure white background, containing only the target object. However, the complex nature of urban environments poses significant challenges in acquiring object instances paired with both registered point clouds and segmented image data, crucial for constructing comprehensive multimodality datasets for completion. Currently, the performance improvement of multimodality methods in this area is limited. Therefore, this paper focuses on exploring point cloud-based methods for completing the shape of buildings.

GAN-based approaches~\cite{xie2021style} leverage the discriminator's implicit learning capability to evaluate the accuracy of complete point clouds produced by the generator. This technique may integrate the previously mentioned encoder-decoder architecture or utilize multimodality data. While capable of generating complete predictions, the outcomes may present multiple interpretations, potentially resulting in point clouds of the same object but with varying styles or details. Given the stringent accuracy requirements for 3D data in disciplines such as geoinformation and geodesy, GAN-based techniques may not always be the optimal choice for completing building point clouds. This is due to the potential for the generated results to diverge from the true geometric configuration of the original structures.

\section{The Building-PCC datasets}\label{sec:method}

In this section, we detail the construction process of the Building-PCC dataset. Our goal is to pair individual building models with their sampling points and corresponding partial point clouds of the actual scanned buildings. In subsection~\ref{sec:data_collect}, we discuss the specifics of the research area and data collection. In subsection~\ref{sec:buidling_extract}, we showcase our approach for extracting complete building instances from both point clouds and models, as well as conducting point cloud sampling.

\subsection{Data collection}\label{sec:data_collect}
We selected The Hague and Rotterdam in the Netherlands as our main research areas, which together host approximately 350,000 buildings of various types. The city of The Hague provides LoD2 building models as open data based on the CityGML standard, from which we collected data for the year 2022. Similarly, the city of Rotterdam also provided LoD2 building models in DWG (from AutoCAD) and SKP (from SketchUp) formats. It is worth mentioning that these building models were semi-automatically constructed by professional modelers using 3D modeling software and GIS data, offering accuracy superior to automatic reconstruction methods. We randomly selected about 50,000 building models from these two cities as our sample, each model containing a unique BAG identifier, the identifier used in the Dutch national geographic database to uniquely identify a building.

For point cloud data, we relied on the Actueel Hoogtebestand Nederland (AHN), a nationwide elevation dataset of the Netherlands. The AHN dataset, collected via LiDAR technology, provides accurate terrain and elevation information. We used data from both AHN3 and AHN4 versions, where AHN3 offers a point density of approximately 8 points per square meter, and AHN4 increases the point density to approximately 10 points per square meter, providing more detailed elevation maps. It is important to note that although AHN3 and AHN4 offer rich information, they do not contain segmentation information for building instances.

\subsection{Building instance extraction}\label{sec:buidling_extract}

Our goal is to extract individual building models for sampling and match them with corresponding real scans. By utilizing the BAG identifiers mentioned in Section~\ref{sec:data_collect}, we are able to precisely extract individual buildings from the LoD2 building models discussed earlier. Using FME software, we achieved this process and converted all models to obj format for sampling. These models may contain holes, inner faces, or might be non-manifold or non-watertight, which could affect the quality and precision of subsequent point cloud sampling. Therefore, we used Easy3d~\cite{nan2021easy3d}, an open-source libraries, to automatically correct these errors. To obtain the ground truth, we employed Poisson Disk Sampling and fixed the number of sampling points at 16,384 per building model. Compared to Uniform Sampling, Poisson Disk Sampling generates a more natural and irregular distribution of sampling points, avoiding the aggregation of sample points and ensuring the randomness and uniformity of sampling.

For AHN point clouds, considering that manually segmenting each building is very time-consuming and automatic segmentation methods based on machine learning require a large amount of ground truth annotation data, we implemented a building instance segmentation pipeline based on LoD2 models (see Figure~\ref{fig:fig2}). First, we used the Overhead Shadow technique to project 3D buildings to extract each building model's footprint. Then, we corrected topological errors such as gaps and overlaps in the footprint and dissolved footprints with common boundaries. Finally, we projected the point cloud to 2D and clipped it using the generated footprint to retain points within the footprint. Thus, we acquired complete point clouds for each building, along with two types of partial point clouds derived from AHN3 and AHN4 datasets, to evaluate the performance of point cloud completion methods.

\section{Benchmarks}\label{sec:benchmarks}

In this section, we embark on an in-depth analysis of the performance of representative deep learning methods applied to the Building-PCC dataset.

\subsection{Implementation details}\label{sec:imple}
In this study, we deployed an experimental setup equipped with an AMD Ryzen Threadripper 1920X 12-core processor, 32GB RAM, and an Nvidia Geforce RTX 3090 graphics card with 24GB of VRAM. All test networks were developed and implemented using the Pytorch framework. We randomly selected approximately 50,000 buildings as the subject of our study, with 30,000 used to construct the training set and 10,000 each allocated to the validation and test sets, supporting model training and performance evaluation. During the training phase, for each building, we randomly choose either an incomplete point cloud from the corresponding AHN3 or AHN4 datasets to serve as the input. Additionally, the point cloud data for each building underwent normalization processing.

\subsection{Representative baselines}\label{sec:baselines}
To comprehensively assess the performance of existing point cloud completion pipelines on real-world dataset, we carefully selected eight representative methods as the solid baseline for the Building-PCC benchmark. These baselines encompass the point cloud-based deep learning strategies discussed in Section 2.2. A concise overview of these baseline methods is as follows:
\begin{itemize}
    \item \textbf{PCN}: As the first architecture in the field of point cloud completion to introduce neural networks for direct processing of 3D coordinates, PCN's encoder transforms the input data into a high-dimensional global feature vector, while its decoder employs the concept of a folding network, simulating the deformation of a two-dimensional plane to map 2D grid points into 3D space~\cite{yuan2018pcn}.
    \item \textbf{FoldingNet}: This end-to-end deep autoencoder was originally designed to tackle the challenges of unsupervised learning on point clouds. Its core mechanism folds a 2D grid onto the underlying surface of a 3D object. For point cloud completion tasks, FoldingNet utilizes an encoder similar to PCN's, but introduces a folding-based decoder~\cite{yang2017foldingnet}.
    \item \textbf{TopNet}: Proposes a decoder based on a tree topology structure, aimed at constructing structured data by generating subsets of the point cloud, facilitating the completion process~\cite{tchapmi2019topnet}.
    \item \textbf{GRNet}: Introduces a 3D grid as an intermediate representation to regularize unordered point clouds, enhancing the structure and contextual information of the point cloud through a gridding residual network to achieve complete cloud reconstruction~\cite{xie2020grnet}.
    \item \textbf{SnowflakeNet}: Simulates the process of completing point clouds as a snowflake-like growth in space, using multi-layer snowflake points for deconvolution to generate locally compact and structured data with high-detail geometries~\cite{xiang2021snowflakenet}.
    \item \textbf{PoinTr}: Transforms the point cloud completion task into a set-to-set translation task, employing a transformer to precisely model the local geometric relations within the structure, addressing the inductive biases in 3D geometry~\cite{yu2021pointr}.
    \item \textbf{AnchorFormer}: Introduces the use of anchors to dynamically capture the regional information of objects. These anchors are dispersed to observed and unobserved positions by estimating specific offsets, forming a sparse set with downsampled points from the input observation data, and finally reconstructing fine-grained 3D structures through a point morphing scheme~\cite{chen2023anchorformer}.
    \item \textbf{AdaPoinTr}: Building on PoinTr, it adds an adaptive query generation mechanism and a denoising module to effectively deal with issues such as discontinuous and uneven appearance~\cite{yu2023adapointr}.
\end{itemize}

These methods showcase the diversity and innovation within the point cloud completion technology, providing a solid foundation for understanding and advancing the field.

\subsection{Evaluation metrics}\label{sec:eval}
Our evaluation methodology is aligned with previous studies~\cite{yu2023adapointr}, employing the mean Chamfer Distance (CD) with the L1-norm (\(CD\)-\( l_1 \)) as the primary metric to quantify the differences between the predicted building point cloud \( P \) and the ground truth \( G \) as defined in Equation~(\ref{equ:1}). Note that a lower \(CD\)-\( l_1 \) value indicates smaller differences between the predicted data and the ground truth, which corresponds to better performance.

\begin{equation}\label{equ:1}
    d_{CD}(P, G) = \frac{1}{|P|} \sum_{p \in P} \min_{g \in G} \| p - g \| + \frac{1}{|G|} \sum_{g \in G} \min_{p \in P} \| g - p \|
\end{equation}

Additionally, we incorporate the F-Score as an alternate evaluation metric. The precision \( P(d) \), recall \( R(d) \), and \(F\)-\(Score(d)\) for a point cloud completion at a given threshold \( d \) are defined as Equations~(\ref{equ:2}) to~(\ref{equ:4}). In our experimental setup, we establish the threshold \( d \) at 1\%.

\begin{equation}\label{equ:2}
    P(d) = \frac{1}{|P|} \sum_{p \in P} \left[ \min_{g \in G} \| p - g \| < d \right]
\end{equation}

\begin{equation}\label{equ:3}
    R(d) = \frac{1}{|G|} \sum_{g \in G} \left[ \min_{p \in P} \| g - p \| < d \right]
\end{equation}

\begin{equation}\label{equ:4}
    F\textrm{-}Score(d) = \frac{2P(d)R(d)}{P(d) + R(d)}
\end{equation}

\subsection{Results}\label{sec:result} 
\begin{figure*}[ht!] %
\begin{adjustwidth}{-3cm}{-3cm}
	\centering
	\begin{tabular}{cm{0.14\textwidth}<{\centering}m{0.14\textwidth}<{\centering}m{0.14\textwidth}<{\centering}m{0.14\textwidth}<{\centering}m{0.14\textwidth}<{\centering}m{0.14\textwidth}<{\centering}}	
            Input &
		\includegraphics[width=0.14\textwidth, height=0.08\textheight, keepaspectratio=true]{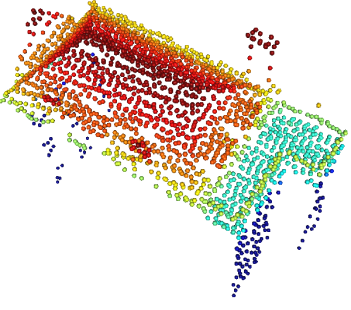}&
		\includegraphics[width=0.14\textwidth, height=0.08\textheight, keepaspectratio=true]{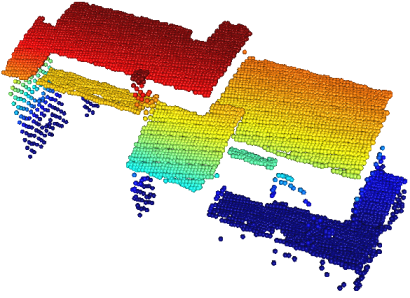}&
		\includegraphics[width=0.14\textwidth, height=0.08\textheight, keepaspectratio=true]{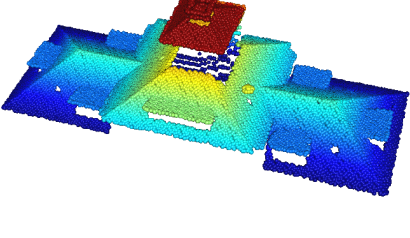}&
		\includegraphics[width=0.14\textwidth, height=0.08\textheight, keepaspectratio=true]{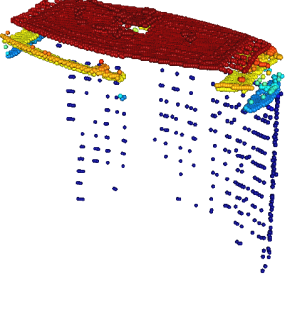}&
  		\includegraphics[width=0.14\textwidth, height=0.08\textheight, keepaspectratio=true]{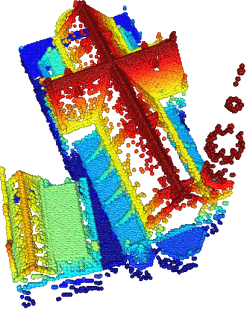}&
		\includegraphics[width=0.14\textwidth, height=0.08\textheight, keepaspectratio=true]{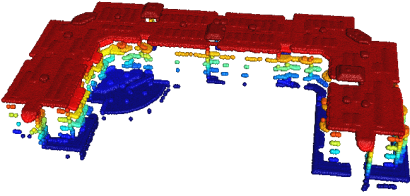}\\	
		PCN &
		\includegraphics[width=0.14\textwidth, height=0.08\textheight, keepaspectratio=true]{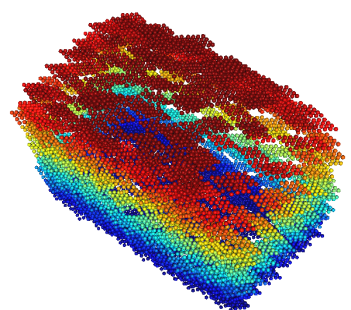}&
		\includegraphics[width=0.14\textwidth, height=0.08\textheight, keepaspectratio=true]{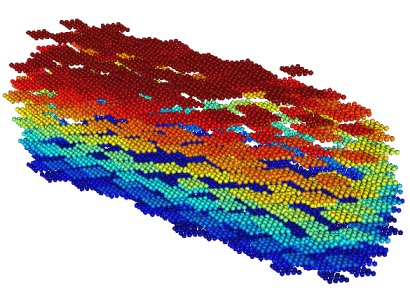}&
		\includegraphics[width=0.14\textwidth, height=0.08\textheight, keepaspectratio=true]{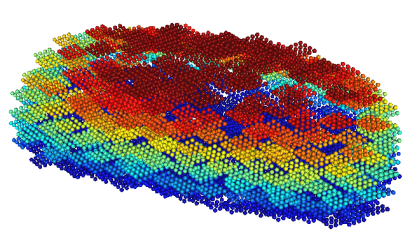}&
		\includegraphics[width=0.14\textwidth, height=0.08\textheight, keepaspectratio=true]{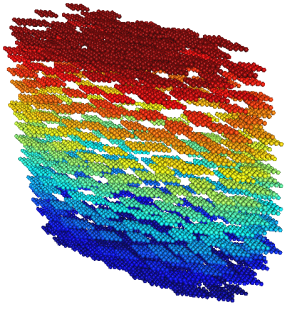}&
  		\includegraphics[width=0.14\textwidth, height=0.08\textheight, keepaspectratio=true]{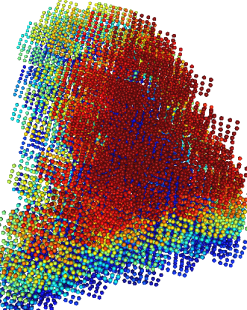}&
		\includegraphics[width=0.14\textwidth, height=0.08\textheight, keepaspectratio=true]{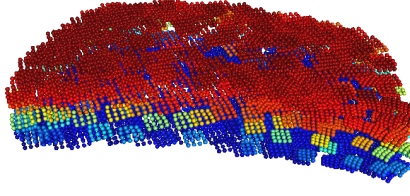}\\
		FoldingNet &
		\includegraphics[width=0.14\textwidth, height=0.08\textheight, keepaspectratio=true]{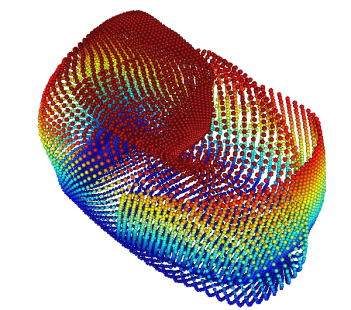}&
		\includegraphics[width=0.14\textwidth, height=0.08\textheight, keepaspectratio=true]{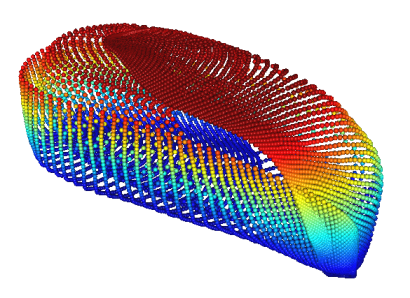}&
		\includegraphics[width=0.14\textwidth, height=0.08\textheight, keepaspectratio=true]{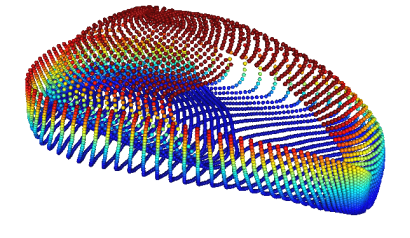}&
		\includegraphics[width=0.14\textwidth, height=0.08\textheight, keepaspectratio=true]{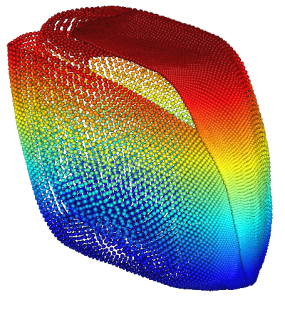}&
  		\includegraphics[width=0.14\textwidth, height=0.08\textheight, keepaspectratio=true]{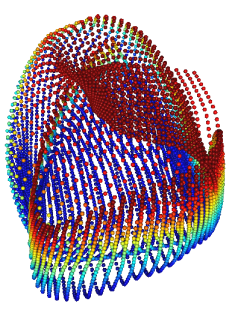}&
		\includegraphics[width=0.14\textwidth, height=0.08\textheight, keepaspectratio=true]{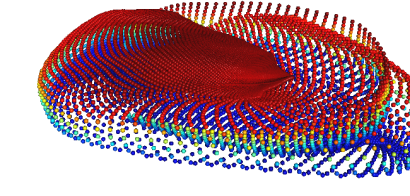}\\	
		TopNet &
		\includegraphics[width=0.14\textwidth, height=0.08\textheight, keepaspectratio=true]{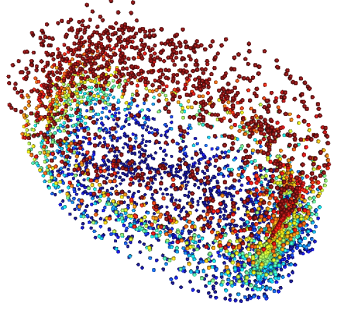}&
		\includegraphics[width=0.14\textwidth, height=0.08\textheight, keepaspectratio=true]{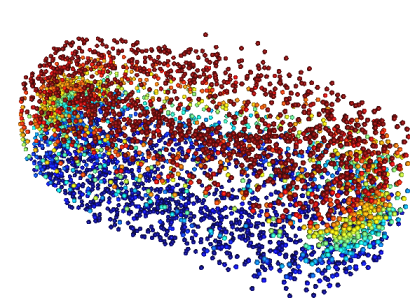}&
		\includegraphics[width=0.14\textwidth, height=0.08\textheight, keepaspectratio=true]{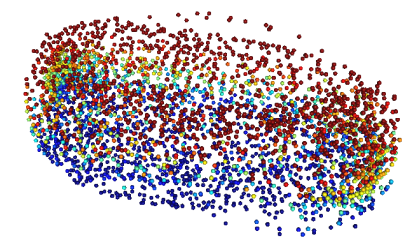}&
		\includegraphics[width=0.14\textwidth, height=0.08\textheight, keepaspectratio=true]{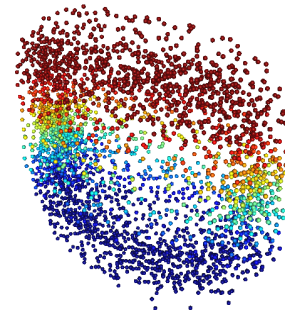}&
  		\includegraphics[width=0.14\textwidth, height=0.08\textheight, keepaspectratio=true]{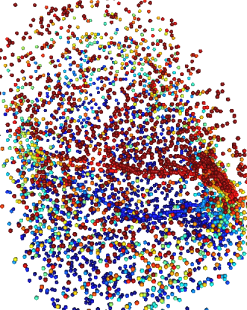}&
		\includegraphics[width=0.14\textwidth, height=0.08\textheight, keepaspectratio=true]{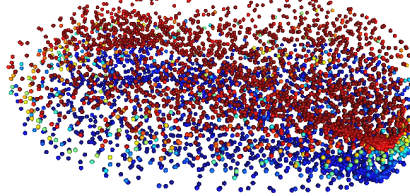}\\
		GRNet &
		\includegraphics[width=0.14\textwidth, height=0.08\textheight, keepaspectratio=true]{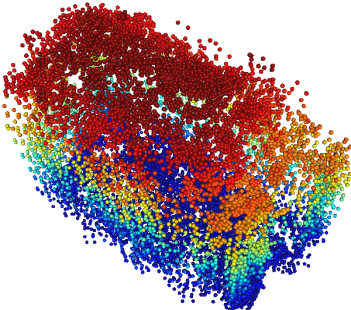}&
		\includegraphics[width=0.14\textwidth, height=0.08\textheight, keepaspectratio=true]{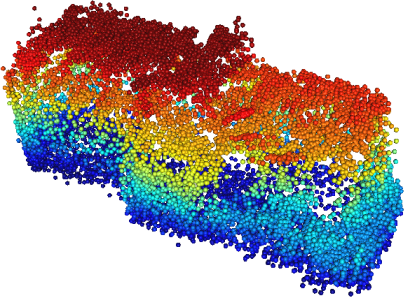}&
		\includegraphics[width=0.14\textwidth, height=0.08\textheight, keepaspectratio=true]{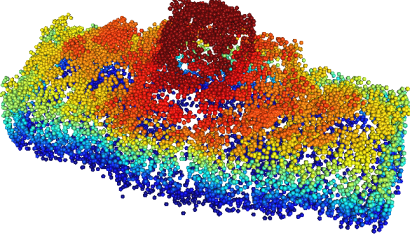}&
		\includegraphics[width=0.14\textwidth, height=0.08\textheight, keepaspectratio=true]{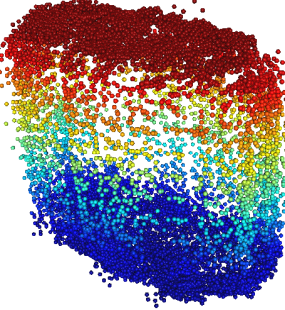}&
  		\includegraphics[width=0.14\textwidth, height=0.08\textheight, keepaspectratio=true]{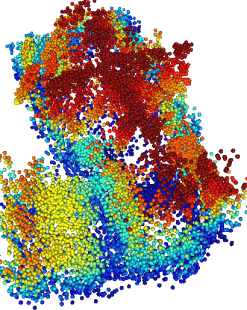}&
		\includegraphics[width=0.14\textwidth, height=0.08\textheight, keepaspectratio=true]{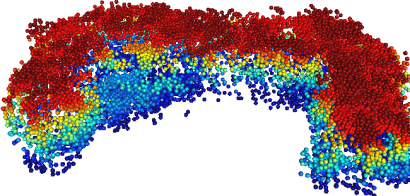}\\
		SnowflakeNet &
		\includegraphics[width=0.14\textwidth, height=0.08\textheight, keepaspectratio=true]{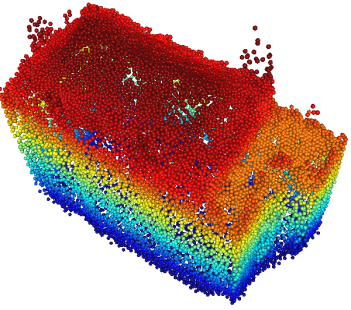}&
		\includegraphics[width=0.14\textwidth, height=0.08\textheight, keepaspectratio=true]{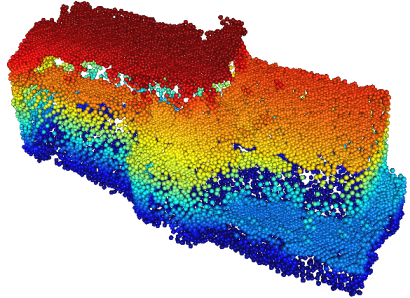}&
		\includegraphics[width=0.14\textwidth, height=0.08\textheight, keepaspectratio=true]{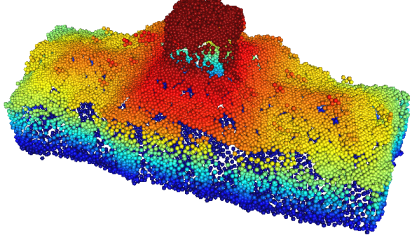}&
		\includegraphics[width=0.14\textwidth, height=0.08\textheight, keepaspectratio=true]{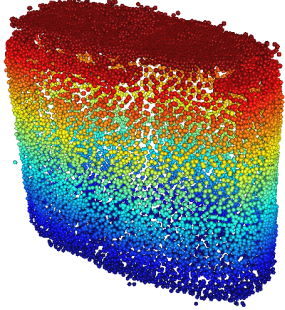}&
  		\includegraphics[width=0.14\textwidth, height=0.08\textheight, keepaspectratio=true]{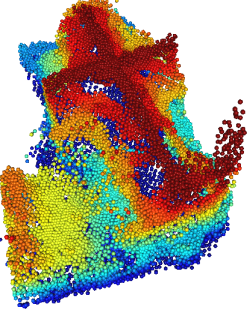}&
		\includegraphics[width=0.14\textwidth, height=0.08\textheight, keepaspectratio=true]{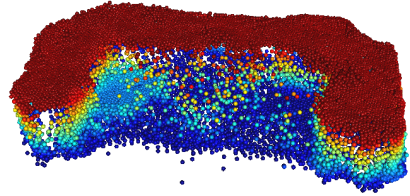}\\
		PoinTr &
		\includegraphics[width=0.14\textwidth, height=0.08\textheight, keepaspectratio=true]{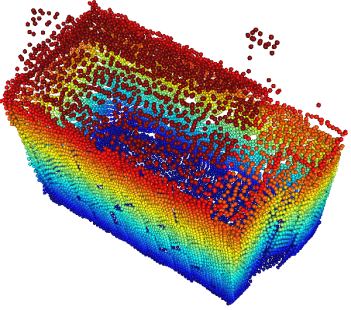}&
		\includegraphics[width=0.14\textwidth, height=0.08\textheight, keepaspectratio=true]{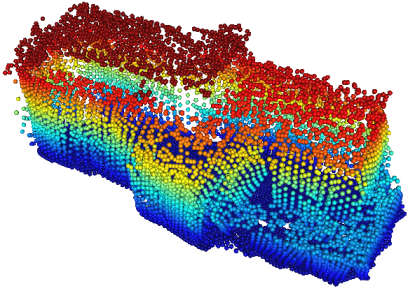}&
		\includegraphics[width=0.14\textwidth, height=0.08\textheight, keepaspectratio=true]{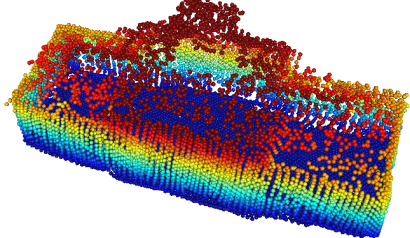}&
		\includegraphics[width=0.14\textwidth, height=0.08\textheight, keepaspectratio=true]{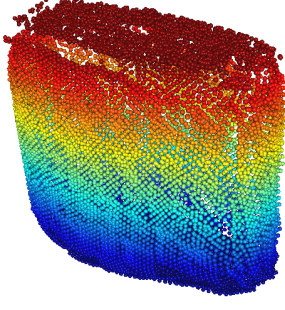}&
  		\includegraphics[width=0.14\textwidth, height=0.08\textheight, keepaspectratio=true]{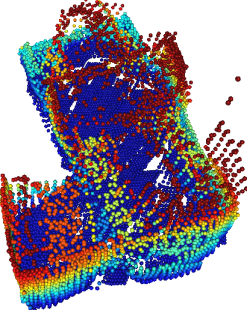}&
		\includegraphics[width=0.14\textwidth, height=0.08\textheight, keepaspectratio=true]{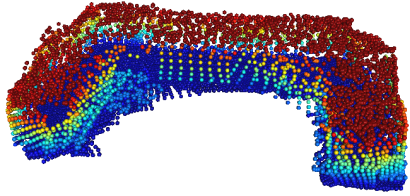}\\
		AnchorFormer &
		\includegraphics[width=0.14\textwidth, height=0.08\textheight, keepaspectratio=true]{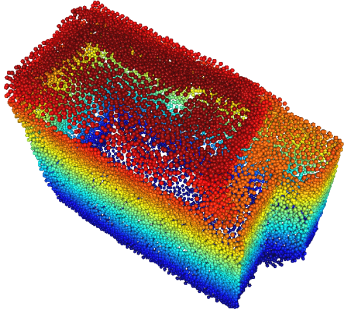}&
		\includegraphics[width=0.14\textwidth, height=0.08\textheight, keepaspectratio=true]{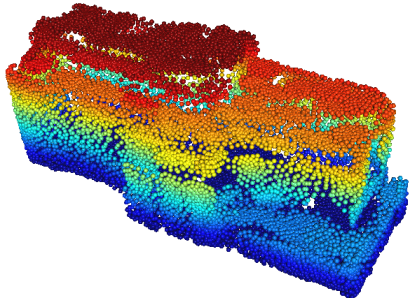}&
		\includegraphics[width=0.14\textwidth, height=0.08\textheight, keepaspectratio=true]{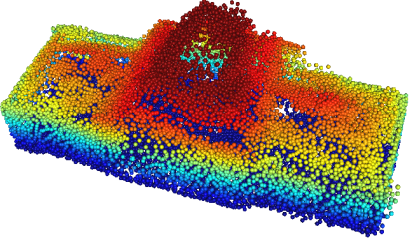}&
		\includegraphics[width=0.14\textwidth, height=0.08\textheight, keepaspectratio=true]{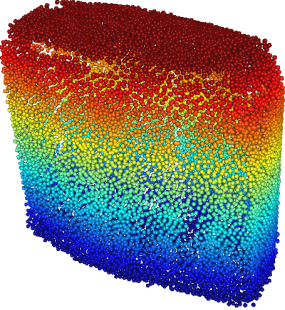}&
  		\includegraphics[width=0.14\textwidth, height=0.08\textheight, keepaspectratio=true]{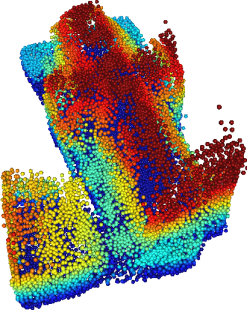}&
		\includegraphics[width=0.14\textwidth, height=0.08\textheight, keepaspectratio=true]{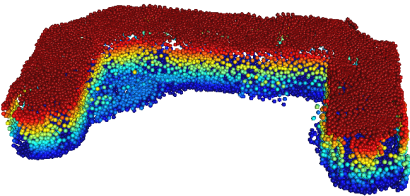}\\
		AdaPoinTr &
		\includegraphics[width=0.14\textwidth, height=0.08\textheight, keepaspectratio=true]{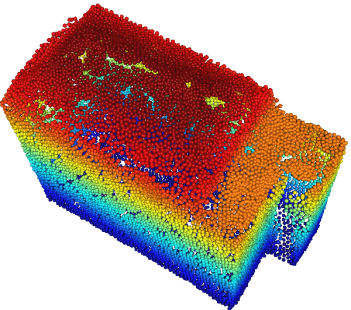}&
		\includegraphics[width=0.14\textwidth, height=0.08\textheight, keepaspectratio=true]{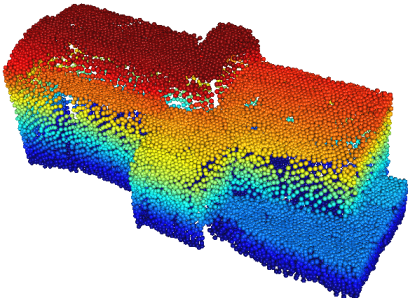}&
		\includegraphics[width=0.14\textwidth, height=0.08\textheight, keepaspectratio=true]{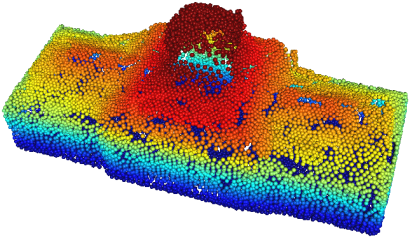}&
		\includegraphics[width=0.14\textwidth, height=0.08\textheight, keepaspectratio=true]{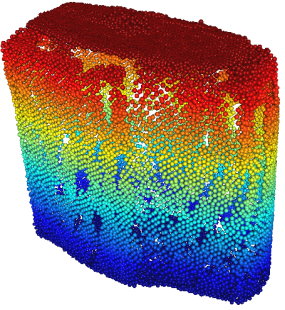}&
  		\includegraphics[width=0.14\textwidth, height=0.08\textheight, keepaspectratio=true]{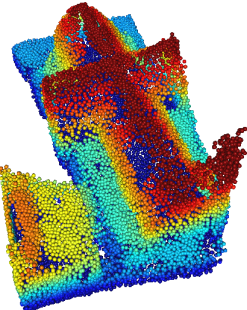}&
		\includegraphics[width=0.14\textwidth, height=0.08\textheight, keepaspectratio=true]{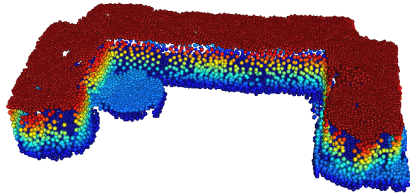}\\
		Ground Truth &
		\includegraphics[width=0.14\textwidth, height=0.08\textheight, keepaspectratio=true]{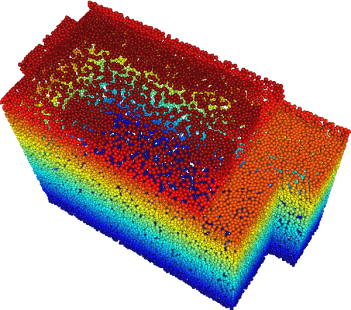}&
		\includegraphics[width=0.14\textwidth, height=0.08\textheight, keepaspectratio=true]{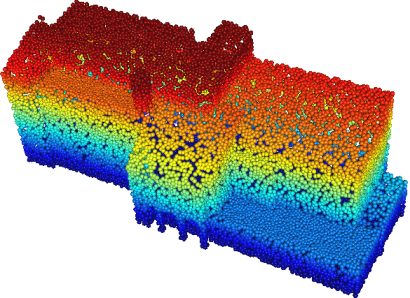}&
		\includegraphics[width=0.14\textwidth, height=0.08\textheight, keepaspectratio=true]{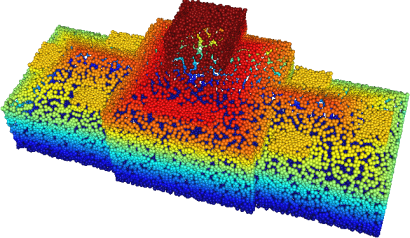}&
		\includegraphics[width=0.14\textwidth, height=0.08\textheight, keepaspectratio=true]{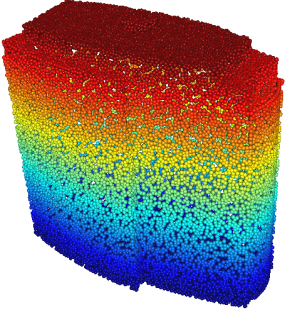}&
  		\includegraphics[width=0.14\textwidth, height=0.08\textheight, keepaspectratio=true]{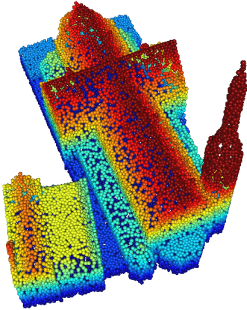}&
		\includegraphics[width=0.14\textwidth, height=0.08\textheight, keepaspectratio=true]{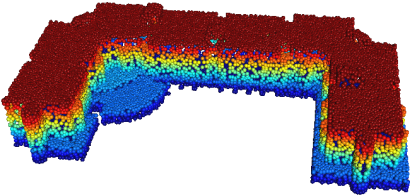}\\
		Model &
		\includegraphics[width=0.14\textwidth, height=0.08\textheight, keepaspectratio=true]{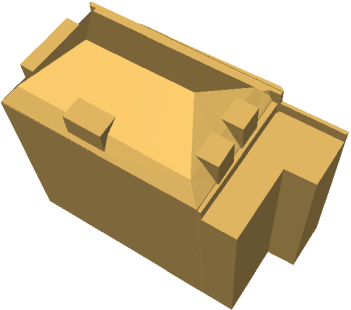}&
		\includegraphics[width=0.14\textwidth, height=0.08\textheight, keepaspectratio=true]{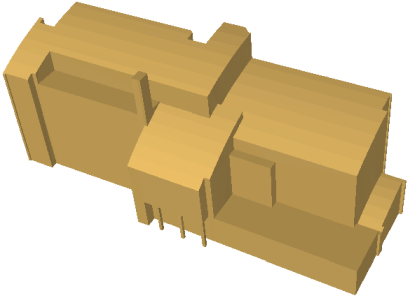}&
		\includegraphics[width=0.14\textwidth, height=0.08\textheight, keepaspectratio=true]{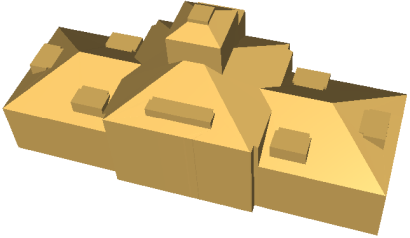}&
		\includegraphics[width=0.14\textwidth, height=0.08\textheight, keepaspectratio=true]{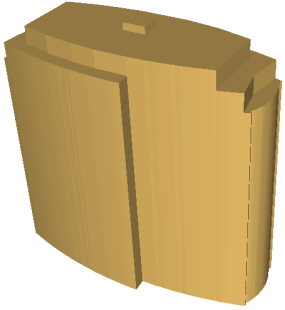}&
  		\includegraphics[width=0.14\textwidth, height=0.08\textheight, keepaspectratio=true]{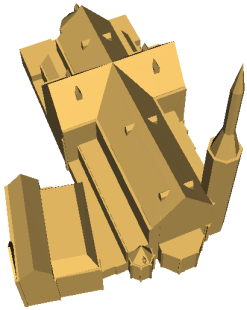}&
		\includegraphics[width=0.14\textwidth, height=0.08\textheight, keepaspectratio=true]{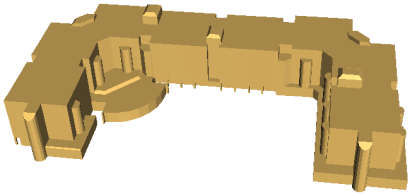}\\
  		& (a) & (b) & (c) & (d) & (e) & (f) \\	
	\end{tabular}
\end{adjustwidth}
\caption{Six visual examples of point cloud completion results by different approaches on the Building-PCC dataset. The color coding, ranging from blue to red, represents the height field, with blue indicating lower elevations and red signifying higher elevations. }
\label{fig:fig3}
\end{figure*}

In this section, we showcase the outcomes of the methods discussed in Section~\ref{sec:baselines} on the Building-PCC dataset, detailed in Table~\ref{tab:tab1} and Figure~\ref{fig:fig3} for quantitative and qualitative analyses, respectively. For the quantitative aspect, 10,000 partial building point clouds from each of the AHN3 and AHN4 datasets served as input for evaluating the overall performance of each method. As delineated in Table~\ref{tab:tab1}'s left section, AnchorFormer, PoinTr, and AdaPoinTr significantly outperform others in terms of average \(CD\)-\( l_1 \), with PoinTr leading at an impressive 1.40. These methods also show robustness across AHN3 and AHN4 inputs, maintaining consistent average Chamfer Distances.

\begin{table*}[ht!]
\centering
\begin{tabularx}{1.0\textwidth}{l|cc|cc|ccccccc}
\toprule
\diagbox{Methods}{AHN3/4} &\(CD\)-\( l_1 \) & \(F\)-\(Score\) & \(CD\)-\( l_1 \) &  \(F\)-\(Score\) & (a) & (b) & (c) & (d) & (e) & (f) & \(CD\)-\( l_1 \) \\
\midrule
PCN          & 6.09 & 0.374 & 6.14 & 0.416 & 14.21 & 14.36 & 14.56 & 11.60 & 11.77 & 12.98 & 13.25 \\
FoldingNet   & 4.20 & 0.328 & 5.63 & 0.352 & 10.31 & 15.47 & 16.03 & 15.60 & 15.04 & 14.98 & 14.57 \\
TopNet       & 6.42 & 0.262 & 6.44 & 0.267 & 14.61 & 12.80 & 12.67 & 11.24 & 11.38 & 15.56 & 13.04 \\
GRNet        & 5.21 & 0.399 & 4.89 & 0.463 & 10.90 & 16.39 & 16.83 & 14.56 & 14.86 & 16.29 & 14.97 \\
SnowflakeNet & 6.61 & 0.588 & 6.60 & 0.629 & 16.24 & 20.92 & 21.08 & 15.88 & 15.91 & 16.53 & 17.76 \\
PoinTr       & \textbf{1.40} & 0.511 & \textbf{1.40} & 0.585 & \textbf{2.81}  & \textbf{5.00}  & 5.21  & \textbf{7.58}  & 7.99  & \textbf{11.39 }& \textbf{6.66 } \\
AnchorFormer & 1.46 & 0.631 & 1.46 & 0.685 & 2.89  & 5.26  & 5.29  & 7.82  & \textbf{7.91}  & 11.70 & 6.81  \\
AdaPoinTr    & 1.42 & \textbf{0.679} & 1.41 & \textbf{0.725} & 2.87  & 5.17  & \textbf{5.18}  & 7.75  & 8.04  & 11.66 & 6.78  \\
\bottomrule
\end{tabularx}
\caption{Comparative evaluation of point cloud completion methods on Building-PCC datasets. The left section quantifies the overall performance across the entire test dataset, with inputs derived separately from AHN3 and AHN4 partial point clouds (with 10,000 samples respectively) , using the mean L1 Chamfer Distance (\(CD\)-\( l_1 \)) and mean \(F\)-\(Score\) metrics. The right section details the performance for selected individual building samples, referenced in Figure~\ref{fig:fig3}, providing specific \(CD\)-\( l_1 \) values for each sample labeled (a) to (f). The last column displays the average \(CD\)-\( l_1 \) across all samples. Optimal performances are denoted in bold for emphasis.}
\label{tab:tab1}
\end{table*}

AdaPoinTr, with its advanced denoising module, excels on the \(F\)-\(Score\) metric, achieving 0.679 and 0.725 on AHN3 and AHN4, respectively, edging out the runner-up, AnchorFormer, by approximately 0.048 and 0.040. Overall, AdaPoinTr stands out in comprehensive performance, closely following PoinTr in average \(CD\)-\( l_1 \) by less than 0.02, yet far surpasses other methods in \(F\)-\(Score\).

To further illustrate these models' capabilities, we conducted tests on six representative buildings for each method, presenting these findings on the right section of Table~\ref{tab:tab1} and in Figure~\ref{fig:fig3}. The right section of Table~\ref{tab:tab1} echoes the quantitative analysis based on \(CD\)-\( l_1 \), affirming that AnchorFormer, PoinTr, and AdaPoinTr maintain superior performance, particularly PoinTr. Figure~\ref{fig:fig3} reveals that while PCN, FoldingNet, and TopNet exhibit notable deviations from the ground truth, the rest successfully reconstructs the overall geometric structure of the buildings. Among these, AdaPoinTr and SnowflakeNet distinctly excel in capturing building details. Nonetheless, the predicted point clouds of all methods exhibit varying degrees of incompleteness in the results and loss of sharp features and fine details compared to the ground truth. These shortcomings might limit their use in downstream applications. Therefore, in Section~\ref{sec:challenge}, we elaborate on the specific reasons for these issues and potential solutions.

\section{Challenges}\label{sec:challenge}

In this section, we delve into challenges faced by point cloud completion methods when dealing with real-world building point clouds. These challenges include imbalanced datasets, limitations of completion structures, and issues of point cloud normalization. Through detailed analysis, we aim to provide insights for researchers on how to effectively utilize point cloud completion methods in practical applications.

\subsection{Impact of imbalanced datasets}\label{sec:imba} 

\begin{figure}[ht!]
\centering
\includegraphics[width=0.3\textwidth]{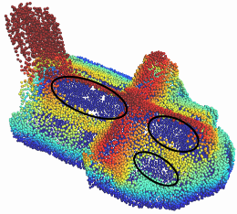}
\caption{AdaPoinTr prediction example using an AHN4 partial point cloud as input (refer to Figure~\ref{fig:fig1b}). The area within the black circle highlights the roof parts that were not successfully predicted. The color coding from blue to red indicates elevation from low to high.}
\label{fig:fig4}
\end{figure}

In classification or semantic segmentation tasks, data imbalance manifests as a discrepancy in sample numbers across categories within the training dataset. This results in models performing better on categories with more samples and worse on underrepresented categories, affecting the overall model performance. A strategy to counter this, is data augmentation to balance sample numbers across categories, preventing model overfitting to dominant data categories. In point cloud completion tasks, while no data imbalance issue from semantic labels exists, an uneven distribution of local geometric structures within point clouds does. Datasets like ShapeNet can augment data by increasing point clouds for objects of the same category in different styles. But for real-world building point clouds, incomplete areas are mostly on the bottoms and facades, not on the roofs. This results in models more easily completing geometric structures of these areas, causing inaccuracies in predicting complete geometric structures of missing roof parts (see Figure~\ref{fig:fig4}).

In response to these issues, we propose two potential research directions. First, we can create synthetic datasets for building point cloud completion, such as using tools like HELIOS++~\cite{winiwarter2022virtual} to simulate laser scans of large-scale virtual urban scenes, including various urban objects like vegetation, street furniture, and building models, to generate building point clouds. By controlling the quantity, location, and attributes of different categories of objects, we can produce a diverse set of incomplete building point clouds. Alternatively, direct simulation scans of LoD2 building models to obtain representative incomplete data can be performed, with random deletion of points to simulate incompleteness. The second approach involves the use of multimodal data, such as orthophotos or multi-view images. Although collecting and matching these data types can be challenging, in theory, the photo-consistency of images can assist in better perceiving the geometric structural consistency.

\subsection{Limitations on fine details}\label{sec:limit_rc} 

In Section~\ref{sec:baselines}, we discussed representative methods, but these generally exhibit loss of details and over-smoothed sharp features when predicting complete building point clouds. These issues often lead to reduced accuracy of the point cloud, thereby affecting the quality of downstream applications such as point cloud-based building model reconstruction.

\begin{figure}[ht!]
\centering
\subfigure[LoD2 building model]{\label{fig:fig5a}\includegraphics[width=0.48\columnwidth]{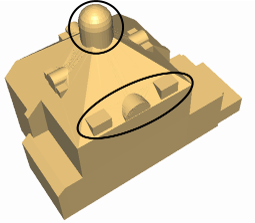}}
\subfigure[AdaPoinTr prediction]{\label{fig:fig5b}\includegraphics[width=0.48\columnwidth]{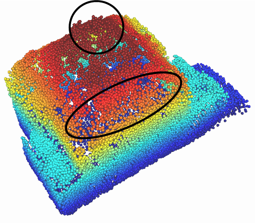}}
\caption{An example of details loss in the predict building point cloud using AdaPoinTr. The black circle emphasizes the absence of detailed roof superstructures.}
\label{fig:fig5}
\end{figure}

Detail loss refers to algorithms' inability to accurately reproduce small building components, like chimneys, dormers, and roof decorations, which are relatively small geometric structures (see Figure~\ref{fig:fig5}). The primary reason is the low resolution of predicted point clouds, i.e., a fixed number of predicted points (such as 16,384 points per building), insufficient to cover large-scale buildings with many details comprehensively. Although increasing the number of predicted points or predicting only missing parts can mitigate this issue, the number of points cannot be indefinitely increased due to memory limitations. Another strategy is to decompose large building point clouds into subsets, predict complete structures for subsets, and then stitch predictions together.

\begin{figure}[ht!]
\centering
\subfigure[AdaPoinTr prediction]{\label{fig:fig6a}\includegraphics[width=0.48\columnwidth]{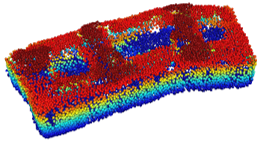}}
\hfill
\subfigure[Ground truth]{\label{fig:fig6b}\includegraphics[width=0.48\columnwidth]{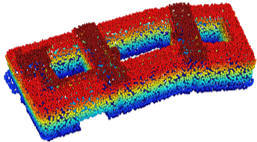}}
\subfigure[AdaPoinTr mesh]{\label{fig:fig6c}\includegraphics[width=0.48\columnwidth]{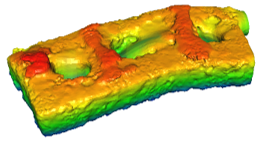}}
\hfill
\subfigure[Ground truth mesh]{\label{fig:fig6d}\includegraphics[width=0.48\columnwidth]{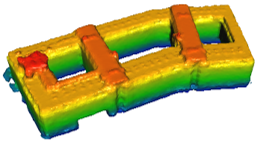}}
\caption{An example of over-smoothed sharp features. The Poisson surface reconstruction is used to extract surface mesh from the point clouds. The color transition from blue to red signifies a shift from lower to higher elevation levels.}
\label{fig:fig6}
\end{figure}

As for the over-smoothed sharp features, they refer to the excessively smoothed sharp corners or sharp edges of buildings. These sharp features primarily arise at the intersections of piece-wise planar surfaces of buildings, such as where the facade meets the roof. As illustrated in Figure~\ref{fig:fig6}, we employed Poisson surface reconstruction on both the AdaPoinTr-predicted point clouds and the ground truth data to highlight the disparities in sharp features. It is evident that the mesh reconstructed from AdaPoinTr-predicted point clouds exhibits more over-smoothed areas compared to the ground truth mesh. The over-smoothed sharp features is primarily due to the fact that the loss functions of these methods are mostly based on the Chamfer Distance, which is a point-to-point distance measure. For urban buildings that conform to the Manhattan world assumption, this measure cannot ensure that the distance error from each point to the local plane is minimized. Therefore, we suggest incorporating the consideration of the distance from points to local planes into the loss function, aiming to address the issue of over-smoothed sharp features to a certain extent.

\subsection{Normalization issues}\label{sec:limit_rc} 

\begin{figure}[ht!]
\centering
\subfigure[]{\label{fig:fig7a}\includegraphics[width=0.32\columnwidth]{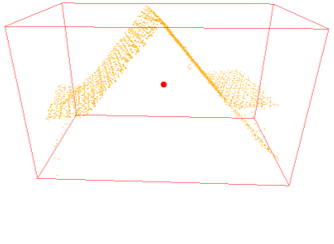}}
\subfigure[]{\label{fig:fig7b}\includegraphics[width=0.32\columnwidth]{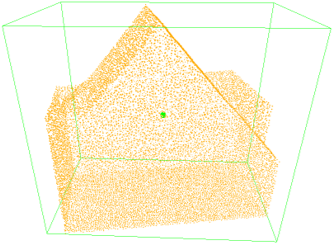}}
\subfigure[]{\label{fig:fig7c}\includegraphics[width=0.32\columnwidth]{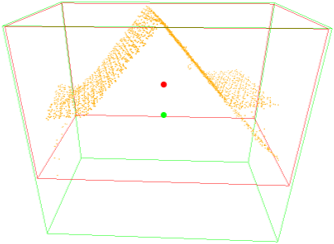}}
\caption{An example of normalization in point cloud completion: (a) displays the partial point cloud with its center point and bounding box outlined in red. (b) presents the ground truth point cloud, marked with a center point and bounding box in green. (c) overlays (a) with the center point and bounding box from (b), emphasizing the disparities in center position and scale between the partial and ground truth point clouds.} 
\label{fig:fig7}
\end{figure}

Although normalization is widely regarded as a means to speed up training, enhance model convergence, and improve generalization capabilities, it has become a problematic issue in the field of point cloud completion. This is because existing methods typically normalize partial point clouds by aligning them with the ground truth, adjusting the partial point clouds based on the center and scale of the ground truth. However, in practical applications, especially for building point cloud completion tasks, this method is not always feasible, as not all buildings have readily available ground truth or models for reference. If the partial point clouds cannot be accurately aligned with the ground truth, it may lead to failures in model training or prediction. Figure~\ref{fig:fig7} illustrates the disparities in center position and scale between the partial point cloud and the ground truth point cloud.

To address this issue, a potential strategy is to develop a network capable of capturing the global features of point clouds, thereby predicting the center position and scale of the point clouds. Alternatively, we could consider relying on external data resources, such as GIS data, to assist in the normalization process. These approaches have the potential to establish a more effective and robust normalization framework for the task of point cloud completion.

\section{Conclusions}\label{sec:conclusion}

This study established the real-world benchmark Building-PCC dataset and evaluated the performance of current deep learning-based point cloud completion technologies in building point cloud completion. The results demonstrate that, despite deep learning methods excelling in capturing local and global geometric features, the quality of results in real-world scenarios is poor and fails to meet downstream applications' needs. By comparing different technologies on the dataset, we revealed the impact of dataset imbalance, limitations on details, and normalization issues on point cloud completion. Finally, we propose solutions to these challenges, including developing networks that capture global features, improving loss functions, and utilizing external data to assist with imbalance and normalization, offering new directions for future research.
 
{
	\begin{spacing}{1.17}
		\normalsize
		\bibliography{Building_PCC} 
	\end{spacing}
}

\end{document}